# Message-Passing Algorithms for Quadratic Programming Formulations of MAP Estimation

**Akshat Kumar**
Department of Computer Science
University of Massachusetts Amherst
akshat@cs.umass.edu

**Shlomo Zilberstein**
Department of Computer Science
University of Massachusetts Amherst
shlomo@cs.umass.edu

## Abstract

Computing *maximum a posteriori* (MAP) estimation in graphical models is an important inference problem with many applications. We present message-passing algorithms for quadratic programming (QP) formulations of MAP estimation for pairwise Markov random fields. In particular, we use the concave-convex procedure (CCCP) to obtain a locally optimal algorithm for the *non-convex* QP formulation. A similar technique is used to derive a globally convergent algorithm for the *convex* QP relaxation of MAP. We also show that a recently developed expectation-maximization (EM) algorithm for the QP formulation of MAP can be derived from the CCCP perspective. Experiments on synthetic and real-world problems confirm that our new approach is competitive with max-product and its variations. Compared with CPLEX, we achieve more than an order-of-magnitude speedup in solving optimally the convex QP relaxation.

## 1 INTRODUCTION

Probabilistic graphical models provide an effective framework for compactly representing probability distributions over high dimensional spaces and performing complex inference using simple local update procedures. In this work, we focus on the class of undirected models called *Markov random fields* (MRFs) [Wainwright and Jordan, 2008]. A common inference problem in this model is to compute the most probable assignment to variables, also called the *maximum a posteriori* (MAP) assignment. MAP estimation is crucial for many practical applications in computer vision and bioinformatics such as protein design [Yanover et al., 2006; Sontag et al., 2008] among others. Computing MAP exactly is NP-hard for general graphs. Thus approximate inference techniques are often used [Wainwright and Jordan, 2008; Sontag et al., 2010].

Recently, several convergent algorithms have been developed for MAP estimation such as tree-reweighted max-product [Wainwright et al., 2002; Kolmogorov, 2006] and max-product LP [Globerson and Jaakkola, 2007; Sontag et al., 2008]. Many of these algorithms are based on the *linear programming* (LP) relaxation of the MAP problem [Wainwright and Jordan, 2008]. A different formulation of MAP is based on *quadratic programming* (QP) [Ravikumar and Lafferty, 2006; Kumar et al., 2009]. The QP formulation is an attractive alternative because it provides a more compact representation of MAP: In a MRF with $n$ variables, $k$ values per variable, and $|E|$ edges, the QP has $O(nk)$ variables whereas the LP has $O(|E|k^2)$ variables. The large size of the LP makes off-the-shelf LP solvers impractical for several real-world problems [Yanover et al., 2006]. Another significant advantage of the QP formulation is that it is *exact*. However, the QP formulation is non-convex, making global optimization hard. To remedy this, Ravikumar and Lafferty [2006] developed a convex QP relaxation of the MAP problem.

Our main contribution is the analysis of the QP formulations of MAP as a *difference of convex functions* (D.C.) problem, which yields efficient, graph-based message-passing algorithms for both the non-convex and convex QP formulations. We use the *concave-convex procedure* (CCCP) to develop the message passing algorithms [Yuille and Rangarajan, 2003]. Motivated by geometric programming [Boyd et al., 2007], we present another QP-based formulation of MAP and solve it using the CCCP technique. The resulting algorithm is shown to be equivalent to a recently developed *expectation-maximization* (EM) algorithm that provides good performance for large MAP problems [Kumar and Zilberstein, 2010]. The CCCP approach, however, is more flexible than EM and makes it easy to incorporate additional constraints that can tighten the convex QP [Kumar et al., 2009]. All the developed CCCP algorithms are guaranteed to converge to a local optimum for non-convex QPs, and to the global optimum for convex QPs. All the algorithms also provide monotonic improvement in the objective.

We experiment on synthetic benchmarks and real-world protein-design problems [Yanover et al., 2006]. Against max-product [Pearl, 1988], CCCP provides significantly better solution quality, sometimes more than 45% for large Ising graphs. On the real-world protein design problems, CCCP achieves near-optimal solution quality for most instances, and is significantly faster than the max-product LP method [Sontag et al., 2008]. Ravikumar and Lafferty [2006] proposed to solve the convex QP relaxation using standard QP solvers. Our message-passing algorithm for this case provides more than an order-of-magnitude speedup against the state-of-the-art QP solver CPLEX.

## 2 QP FORMULATION OF MAP

A pairwise Markov random field (MRF) is described using an undirected graph $G = (V, E)$. A discrete random variable $x_i$ with a finite domain is associated with each node $i \in V$ of the graph. Associated with each edge $(i, j) \in E$ is a potential function $\theta_{ij}(x_i, x_j)$. The complete assignment $\mathbf{x}$ has the probability:

$$p(\mathbf{x}; \theta) \propto \exp\left(\sum_{ij \in E} \theta_{ij}(x_i, x_j)\right)$$

The MAP problem consists of finding the most probable assignment to all the variables under $p(\mathbf{x}; \theta)$. This is equivalent to finding the assignment $\mathbf{x}$ that maximizes the function $f(\mathbf{x}; \theta) = \sum_{ij \in E} \theta_{ij}(x_i, x_j)$. We assume w.l.o.g. that each $\theta_{ij}$ is nonnegative, otherwise a constant can be added to each $\theta_{ij}$ without changing the optimal solution. Let $p_i$ be the marginal probability associated with each MRF node $i \in V$. The MAP quadratic programming (QP) formulation [Ravikumar and Lafferty, 2006] is given by:

$$\max_{\mathbf{P}} \sum_{ij \in E} \sum_{x_i, x_j} p_i(x_i) p_j(x_j) \theta_{ij}(x_i, x_j) \quad (1)$$

$$\text{subject to } \sum_{x_i} p_i(x_i) = 1, p_i(x_i) \geq 0 \; \forall i \in V$$

The above QP is compact even for large graphical models and has simple linear constraints: $O(nk)$ variables and $n$ normalization constraints where $n = |V|$ and $k$ is the domain size. Ravikumar and Lafferty [2006] also show that this formulation is *exact*. That is, the global optimum of the above QP will maximize the function $f(\mathbf{x}; \theta)$ and an integral MAP assignment can be extracted from it. However this formulation is non-convex, making global optimization hard.

Nonetheless, for several problems, a local optimum of this QP provides a good solution as we will show empirically. This was also observed by Kumar and Zilberstein [2010].

### 2.1 The Concave Convex Procedure

The *concave-convex procedure* (CCCP) [Yuille and Rangarajan, 2003] is a popular approach to optimize a general non-convex function expressed as a *difference* of two convex functions. We use this method to obtain message-passing algorithms for QP formulations of MAP. We describe it here briefly.
Consider the optimization problem:

$$\min\{g(x) : x \in \Omega\} \quad (2)$$

where $g(x) = u(x) - v(x)$ is an arbitrary function with $u$, $v$ being real-valued *convex* functions and $\Omega$ being a convex set. The CCCP method provides an iterative procedure that generates a sequence of points $x^l$ by solving the following convex program:

$$x^{l+1} = \arg\min\{u(x) - x^T \nabla v(x^l) : x \in \Omega\} \quad (3)$$

Each iteration of CCCP decreases the objective $g(x)$ and is guaranteed to converge to a local optimum [Sriperumbudur and Lanckriet, 2009].

### 2.2 Solving MAP QP Using CCCP

We first show how the CCCP framework can be used to solve the QP in Eq. (1). We adopt the convention that a MAP QP always refers to the QP in Eq. (1); the convex variant of this QP shall be explicitly differentiated when addressed later. Consider the following functions $u$, $v$:

$$u(p) = \sum_{ij} \sum_{x_i x_j} \frac{\theta_{ij}(x_i, x_j)}{2} \left(p_i^2(x_i) + p_j^2(x_j)\right) \quad (4)$$

$$v(p) = \sum_{ij} \sum_{x_i x_j} \frac{\theta_{ij}(x_i, x_j)}{2} \left(p_i(x_i) + p_j(x_j)\right)^2 \quad (5)$$

The above functions are convex because the quadratic functions $f(z) = z^2$ and $f(y, z) = (y + z)^2$ are convex, and the nonnegative weighted sum of convex functions is also convex [Boyd and Vandenberghe, 2004, Ch. 3]. It can be easily verified that the QP in Eq. (1) can be written as $\min_p \{u(p) - v(p)\}$ with normalization and nonnegativity constraints defining the constraint set $\Omega$. Intuitively, we used the simple identity $-2xy = (x^2 + y^2) - (x + y)^2$. We also negated the objective function to convert maximization to minimization. For simplicity, we denote the gradient $\partial v/\partial p(x_i)$ by $\nabla_{x_i} v$.

$$\nabla_{x_i} v = p_i(x_i) \sum_{j \in \text{Ne}(i)} \sum_{x_j} \theta_{ij}(x_i, x_j) + \sum_{j \in \text{Ne}(i)} \sum_{x_j} \theta_{ij}(x_i, x_j) p_j(x_j)$$

The first part of the above equation involves a local computation associated with an MRF node and the second part defines the messages $\delta_j$ from neighbors $j \in Ne(i)$ of node $i$. It can be made explicit as follows:

$$\hat{\theta}(x_i) = \sum_{j \in Ne(i)} \sum_{x_j} \theta_{ij}(x_i, x_j); \delta_j(x_i) = \sum_{x_j} \theta_{ij}(x_i, x_j) p_j(x_j)$$

$$\nabla_{x_i} v = p_i(x_i) \hat{\theta}(x_i) + \sum_{j \in Ne(i)} \delta_j(x_i) \qquad (6)$$

**CCCP iterations:** Each iteration of CCCP involves solving the convex program of Eq. (3). First we write the Lagrangian function involving only the normalization constraints, later we address the nonnegativity inequality constraints. $\nabla^l v$ denotes the gradient from the previous iteration $l$.

$$L(p, \lambda) = \sum_{ij} \sum_{x_i x_j} \frac{\theta_{ij}(x_i, x_j)}{2} \{p_i^2(x_i) + p_j^2(x_j)\}$$
$$- \sum_i \sum_{x_i} p_i(x_i) \nabla^l_{x_i} v + \sum_i \lambda_i (\sum_{x_i} p_i(x_i) - 1) \qquad (7)$$

Solving for the first order optimality conditions $\nabla_p L(x^\star, \lambda^\star)$ and $\nabla_\lambda L(x^\star, \lambda^\star)$, we get the solution:

$$p_i^{l+1}(x_i) = \frac{\nabla^l_{x_i} v - \lambda_i}{\hat{\theta}(x_i)} \qquad (8)$$

$$\lambda_i = \frac{1}{\sum_{x_i} \frac{1}{\hat{\theta}(x_i)}} \left( \sum_{x_i} \frac{\nabla^l_{x_i} v}{\hat{\theta}(x_i)} - 1 \right) \qquad (9)$$

**Nonnegativity constraints:** Nonnegativity constraints in the MAP QP are inequality constraints which are harder to handle as the Karush-Kuhn-Tucker (KKT) conditions include the nonlinear complementary slackness condition $\mu_j^\star p_j^\star(x_j) = 0$ [Boyd and Vandenberghe, 2004]. We can use interior-point methods, but they lose the efficiency of graph based message passing. Fortunately, we show that for MAP QP, the KKT conditions are easily satisfied by incorporating an inner-loop in the CCCP iterations.

Alg. 1 shows the complete message-passing procedure to solve MAP QPs. Each outer loop corresponds to solving the CCCP iteration of Eq. (3) and is run until the desired number of iterations is reached. The messages $\delta$ are used for computing the gradient $\nabla v$ as in Eq. (6). The inner loop corresponds to satisfying the KKT conditions including the nonnegativity inequality constraints. Intuitively, the strategy to handle inequality constraints is as highlighted in [Bertsekas, 1999, Sec. 3.3.1] – considering all possible combinations of inequality constraints being active ($p_i(x_i)=0$) or inactive ($p_i(x_i) > 0$) and solving the resulting KKT conditions, which is easier as they become linear equations. If the resulting solution satisfies the KKT conditions of the original problem, then we have a valid solution for the original optimization problem. Of course,

---

**1: Graph-based message passing for MAP estimation**

**input**: Graph $G = (V, E)$ and potentials $\theta_{ij}$ per edge
//outer loop starts
**repeat**
    **foreach** *node* $i \in V$ **do**
        $\delta_{i \to j}(x_j) \leftarrow \sum_{x_i} p_i(x_i) \theta_{ij}(x_i, x_j)$
        Send message $\delta_{i \to j}$ to each neighbor $j \in Ne(i)$
    **foreach** *node* $i \in V$ **do**
        zeros $\leftarrow \phi$
        //inner loop starts
        **repeat**
            Set $p_i(x_i) \leftarrow 0 \;\forall x_i \in$ zeros
            Calculate $p_i(x_i)$ using Eq. (8) $\forall\; x_i \notin$ zeros
            zeros $\leftarrow$ zeros $\cup \{x_i : p_i(x_i) < 0\}$
        **until** *all beliefs* $p_i(x_i) \geq 0$
**until** *stopping criterion is satisfied*
**return**: The decoded complete integral assignment

---

this is highly inefficient for the general case. But fortunately for the MAP QP, we show that the inner loop of Alg. 1 recovers the correct solution and the Lagrange multipliers are computed efficiently for the convex program of Eq. (3). We describe it below.

The inner loop includes local computation to each MRF node $i$ and does not require message passing. Intuitively, the set zeros tracks all the settings $x_i'$ of the variable $x_i$ for which $p_i(x_i')$ was negative in any previous inner loop iteration. It then clamps all such beliefs to 0 for *all* future iterations. Then the beliefs for the rest of the settings of $x_i$ are computed using Eq. (8). The new Lagrange multiplier $\lambda_i$ (which corresponds to the condition $\nabla_\lambda L(x^\star, \lambda^\star) = 0$) is calculated using the equation $\sum_{x_i \setminus x_i'} p_i(x_i) = 1$.

**Lemma 1.** *The inner loop of Alg. 1 terminates with worst case complexity $O(k^2)$, and yields a feasible point for the convex program of Eq. (3).*

*Proof.* The size of the set zeros increases with each iteration, therefore the inner loop must terminate as each variable's domain is finite. With the domain size of a variable being $k$, the inner loop can run for at most $k$ iterations. Computing new beliefs within each inner loop iteration also requires $O(k)$ time. Thus the worst case total complexity is $O(k^2)$.

The inner loop can terminate only in two ways – before iteration $k$ or at iteration $k$. If it terminates before iteration $k$, then it implies that all the beliefs $p_i(x_i)$ must be nonnegative. The normalization constraints are always enforced by the Lagrange multipliers $\lambda_i$'s. If it terminates during iteration $k$, then it implies that $k-1$ settings of the variable $x_i$ are clamped to zero as the size of the set zeros will be *exactly* $k-1$. The size cannot be $k$ because that would make all the beliefs equal to zero, making it impossible to satisfy the normalization constraint; $\lambda_i$ will not allow this. The size cannot be smaller than $k-1$ because the set `zeros` grows by at least one element during each previous iteration. Therefore the only solution during iteration $k$ is to set the single remaining setting of the variable $x_i$ to 1 to satisfy the normalization and nonnegativity constraints simultaneously. Therefore the inner loop always yields a feasible point upon termination. □

Empirically, we observed that even for large protein design problems with $k=150$, the number of required inner loop iterations is below 20 – far below the worst case complexity. For a fixed outer loop $l$, let the inner loop iterations be indexed by $r$.

**Lemma 2.** *The Lagrange multiplier corresponding to the normalization constraint for a MRF variable $x_i$ always increases with each inner loop iteration.*

*Proof.* Each inner loop iteration $r$ computes a new Lagrange multiplier $\lambda_i^r$ for the constraint $\sum_i p_i(x_i) = 1$ using Eq. (8). We show that $\lambda_i^{r+1} > \lambda_i^r$. For the inner loop iteration $r$, some of the computed beliefs must be negative, otherwise the inner loop must have terminated. Let $x_i'$ denote those settings of variable $x_i$ for which $p_i(x_i') < 0$ in iteration $r$. From the normalization constraint for iteration $r$, we get:

$$\sum_{x_i} \frac{\nabla_{x_i}^l v - \lambda_i^r}{\hat{\theta}(x_i)} = 1 \qquad (10)$$

We used the explicit representation of $p_i(x_i)$ from Eq. (8). Since $p_i(x_i')$ are negative, we get:

$$\sum_{x_i \setminus x_i'} \frac{\nabla_{x_i}^l v - \lambda_i^r}{\hat{\theta}(x_i)} > 1 \qquad (11)$$

The belief for all such $x_i'$ will become zero for the next inner loop iteration $r+1$. From the normalization constraint for iteration $r+1$, we get:

$$\sum_{x_i \setminus x_i'} \frac{\nabla_{x_i}^l v - \lambda_i^{r+1}}{\hat{\theta}(x_i)} = 1 \qquad (12)$$

We used a slight simplification in the above equations as we ignored the effect of previous iterations, before iteration $r$. However, it will not change the conclusion as all the beliefs that were clamped to zero earlier (before iteration $r$) shall remain so for all future iterations. Note that $\nabla_{x_i}$ and $\hat{\theta}$ do not depend on the inner loop iterations. Subtracting Eq. 12 from Eq. 11:

$$(\lambda_i^{r+1} - \lambda_i^r) \sum_{x_i \setminus x_i'} \frac{1}{\hat{\theta}(x_i)} > 0 \qquad (13)$$

Since we assumed that all potential functions $\theta_{ij}$ are nonnegative, we must have $(\lambda_i^{r+1} - \lambda_i^r) > 0$. Hence $\lambda_i^{r+1} > \lambda_i^r$ and the lemma is proved. □

**Theorem 3.** *The inner loop of Alg. 1 correctly recovers all the Lagrange multipliers for the equality and inequality constraints for the convex program of Eq. (3), thus solving it exactly.*

*Proof.* Lemma 1 shows that the inner loop provides a feasible point of the convex program. We now show that this point also satisfies the KKT conditions, thus is the optimal solution. The KKT conditions for the normalization constraints are always satisfied during the belief updates (see. Eq. (8)). The main task is to show that for the inequality constraint $-p_i(x_i) \leq 0$, the KKT conditions hold. That is, if $p_i(x_i) = 0$, then the Lagrange multiplier $\mu_i(x_i) \geq 0$, and if $-p_i(x_i) < 0$, then $\mu_i(x_i) = 0$.

By using the KKT condition $\nabla_{p_i(x_i)} L(p^\star, \lambda^\star, \mu^\star) = 0$, we get:

$$\sum_{j \in Ne(i)} \sum_{x_j} \theta_{ij}(x_i, x_j) p_i(x_i) - \nabla_{x_i}^l v + \lambda_i - \mu_i(x_i) = 0 \qquad (14)$$

The main focus of the proof is on the beliefs for elements in the set `zeros`. Let us focus on the end of an inner loop iteration $r$, when a new element $x_i'$ is added to `zeros` because its computed belief $p_i(x_i') < 0$. Using Eq. (8), we know that $p_i(x_i') = \frac{\nabla_{x_i'}^l v - \lambda_i^r}{\hat{\theta}(x_i')}$. Because $p_i(x_i') < 0$ we get:

$$\lambda_i^r > \nabla_{x_i'}^l v \qquad (15)$$

For all future iterations of the inner loop, $p_i(x_i')$ will be set to zero. Therefore the KKT condition for iteration $r+1$ mandates that $\mu_i^{r+1}(x_i') \geq 0$. Setting $p_i(x_i') = 0$ in Eq. (14), we get:

$$\mu_i^{r+1}(x_i') = \lambda_i^{r+1} - \nabla_{x_i'}^l v \qquad (16)$$

We know from Lemma 2 that $\lambda_i^{r+1} > \lambda_i^r$. Combining this fact with Eq. (15), we get $\mu_i^{r+1}(x_i') > 0$, thereby satisfying the KKT condition. Note that the only component depending on the inner loop in the above condition is $\lambda_i^r$; $\nabla_{x_i'}$ is fixed during each inner loop. Furthermore, for all future inner loop iterations, the KKT conditions for all elements $x_i'$'s in the set `zeros` will be met due to the increasing nature of the multiplier $\lambda_i$.

Therefore, when the inner loop terminates, we shall have correct Lagrange multipliers $\mu$ satisfying $\mu \geq 0$ for all the elements of the set `zeros`. For the rest of the elements, the multiplier $\mu = 0$, satisfying all the KKT conditions. As the first order KKT conditions are both necessary and sufficient for optimality in convex programs [Bertsekas, 1999, Sec. 3.3.4], the inner loop solves *exactly* the convex program in Eq. (3). □

## 2.3 Solving Convex MAP QP Using CCCP

Because the previous QP formulation of MAP is nonconvex, global optimization is hard. To remedy this, Ravikumar and Lafferty [2006] developed a convex QP relaxation for MAP, which performed well on their benchmarks. Recently, Kumar et al. [2009] showed that the convex QP relaxation is also equivalent to the second order cone programming (SOCP) relaxation. Ravikumar and Lafferty [2006] proposed to solve such QP using standard QP solvers. We show using CCCP that this QP relaxation can be solved efficiently using graph-based message passing, and the resulting algorithm converges to the global optimum of the relaxed QP. Experimentally, we found the resulting message-passing algorithm to be highly efficient even for large graphs, outperforming CPLEX by more than an order-of-magnitude. The relaxed QP is described as follows:

$$\max_p \sum_i \sum_{x_i} p_i(x_i) d_i(x_i) + \sum_{ij} \sum_{x_i,x_j} p_i(x_i) p_j(x_j) \theta_{ij}(x_i,x_j) - \sum_i \sum_{x_i} p_i^2(x_i) d_i(x_i) \quad (17)$$

The relaxation is based on adding a diagonal term, $d_i(x_i)$, for each variable $x_i$. Note that under the integrality assumption $p_i(x_i) = p_i^2(x_i)$, thus the first and last terms cancel out, resulting in the original MAP QP. The diagonal term is given by:

$$d_i(x_i) = \sum_{j \in Ne(i)} \sum_{x_j} \frac{|\theta_{ij}(x_i,x_j)|}{2}$$

Consider the convex function $u(p)$ represented as:

$$\sum_{ij} \sum_{x_i x_j} \frac{\theta_{ij}(x_i,x_j)}{2} \left( p_i^2(x_i) + p_j^2(x_j) \right) + \sum_{i,x_i} p_i^2(x_i) d_i(x_i)$$

and the convex function $v(p)$ represented as:

$$\sum_{ij} \sum_{x_i x_j} \frac{\theta_{ij}(x_i,x_j)}{2} \left( p_i(x_i) + p_j(x_j) \right)^2 + \sum_{i,x_i} p_i(x_i) d_i(x_i)$$

The above two functions are the same as the original QP formulation, except for the added diagonal terms $d_i(x_i)$. It can be easily verified that the relaxed QP objective can be written as $\min_p \{u(p) - v(p)\}$ subject to normalization and nonnegativity constraints. Note that the maximization of the relaxed QP is converted to minimization by negating the objective. The gradient required by CCCP is given by:

$$\nabla_{x_i} v = p_i(x_i) \hat{\theta}(x_i) + \sum_{j \in Ne(i)} \sum_{x_j} \theta_{ij}(x_i,x_j) p_j(x_j) + d_i(x_i)$$

Notice the close similarity with the MAP QP case in Eq. (6). The only additional term is $d_i(x_i)$, which needs to be computed only once before message passing begins. The messages for the relaxed QP case are exactly the same as the $\delta$ messages for the MAP QP. The Lagrangian corresponding to the convex program of Eq. (3) is similar to the MAP QP case (see Eq. (7)) with an additional term $\sum_{i,x_i} p_i^2(x_i) d_i(x_i)$. The constraint set $\Omega$ includes the normalization and nonnegativity constraints as for the MAP QP case.

Solving for the optimality conditions $\nabla_p L(p^\star, \lambda^\star)$ and $\nabla_\lambda L(p^\star, \lambda^\star)$, we get the new beliefs as follows:

$$p_i^{l+1}(x_i) = \frac{\nabla_{x_i}^l v - \lambda_i}{2d_i(x_i) + \hat{\theta}(x_i)} \quad (18)$$

The Lagrange multiplier $\lambda_i$ for the normalization constraint can be calculated by using the equation $\sum_{x_i} p_i(x_i) = 1$. The only difference from the corresponding Eq. (8) for the MAP QP is the additional term $d_i(x_i)$ in the denominator.

Thanks to these strong similarities, we can show that Alg. 1 also works for the convex MAP QP with minor modifications. First, we calculate the diagonal terms $d_i(x_i)$ once for each variable $x_i$ of the MRF. The message-passing procedure for each outer loop iteration remains the same. The second difference lies in the inner loop that enforces the nonnegativity constraints. The inner loop now uses Eq. (18) instead of Eq. (8) to estimate new beliefs $p_i(x_i)$. The proof is omitted being very similar to the MAP QP case.

**Theorem 4.** *The CCCP message-passing algorithm converges to a global optimum of the convex MAP QP.*

The result is based on the fact that CCCP converges to a stationary point of the given constrained optimization problem that satisfies the KKT conditions [Sriperumbudur and Lanckriet, 2009]. Because the KKT conditions are both necessary and sufficient for convex optimization problems with linear constraints [Bertsekas, 1999], the result follows. We also highlight that a global optimum of the convex QP may not solve the MAP problem exactly, as the convex QP is a variational approximation to MAP that may not be tight. Nonetheless, it has shown to perform well in practice [Ravikumar and Lafferty, 2006].

## 2.4 GP Based Reformulation of MAP QP

We now present another formulation of the MAP QP problem based on *geometric programming* (GP). A GP is a type of mathematical program characterized by objectives and constraints that have a special form. For details, we refer to [Boyd et al., 2007]. While the QP formulation of MAP in Eq. (1) is not exactly a GP, it bears a close resemblance. This allows us to transfer some ideas from GP, which we shall describe. We start with some basic concepts of GP.

**Definition 1.** Let $x_1, \ldots, x_n$ denote real positive variables. A real valued **monomial** function has the form $f(x) = cx_1^{a_1} x_2^{a_2} \cdots x_n^{a_n}$, where $c > 0$ and $a_i \in \Re$. A **posynomial** function is a sum of one or more monomials: $f(x) = \sum_{k=1}^{K} c_k x_1^{a_{1k}} x_2^{a_{2k}} \cdots x_n^{a_{nk}}$

In a GP, the objective function and all inequality constraints are posynomial functions. The MAP QP (see Eq. (1)) satisfies this requirement – the potential function $\theta_{ij}$ corresponds to $c_k$ and is positive (the $\theta_{ij} = 0$ case can be excluded for convenience); node marginals $p_i(x_i)$ are real positive variables. Since we already assumed marginals $p_i$ to be positive, nonnegativity inequality constraints are not required. In a GP, the equality constraints can only be *monomial* functions. This is not satisfied in MAP QP as normalization constraints are posynomials. Nonetheless, we proceed as in GP by converting the original problem using certain optimality-preserving transformations. The first change is to let $p_i(x_i) = e^{y_i(x_i)}$, where $y_i(x_i)$ is an unconstrained variable. This is allowed as all marginals must be positive. The second change is to take the log of the objective function; because log is a monotonic function, this will not change the optimal solution. The reformulated MAP QP is shown below:

$$\min: -\log\left\{\sum_{ij}\sum_{x_i,x_j}\exp\left(y_i(x_i)+y_j(x_j)+\log\theta_{ij}(x_i,x_j)\right)\right\}$$

subject to: $\sum_{x_i} e^{y_i(x_i)} = 1 \;\forall i \in V$

This nonlinear program has the same optimal solutions as the original MAP QP. As log-sum-exp is convex, the objective function of the above problem is concave. Note that we are minimizing a concave function that can have multiple local optima. We again solve it using CCCP. Consider the function $u(y) = 0$ and $v(y)$ as the objective of the above program, but without the negative sign. The gradient required by CCCP is:

$$\nabla_{y_i}^l v = \frac{\sum_{j \in Ne(i)}\sum_{x_j}\theta_{ij}(x_i,x_j)\exp(y_i(x_i)+y_j(x_j))}{\sum_{ij}\sum_{x_i,x_j}\theta_{ij}(x_i,x_j)\exp(y_i(x_i)+y_j(x_j))} \quad (19)$$

The Lagrangian corresponding to Eq. (3) with constraint set including only normalization constraints is:

$$L(y,\lambda) = -\sum_{i,x_i} y_i(x_i)\nabla_{y_i}^l v + \sum_i \lambda_i(\sum_{x_i} e^{y_i(x_i)} - 1)$$

Using the first order optimality condition, we get:

$$\exp\left(y_i(x_i)\right) = \frac{\nabla_{y_i}^l v}{\lambda_i} \quad (20)$$

We note that the denominator of Eq. (19) is a constant for each $y_i(x_i)$. Therefore we represent it using $c^l$. Re-substituting $p_i(x_i) = e^{y_i(x_i)}$ and $\nabla_{y_i}^l v$ in Eq. (20):

$$p_i^\star(x_i) = \frac{\sum_{j \in Ne(i)}\sum_{x_j}\theta_{ij}(x_i,x_j)p_i(x_i)p_j(x_j)}{c^l \lambda_i}$$

where $p_i^\star(x_i)$ is the new parameter for the current iteration, and parameters without asterisk (on the R.H.S.) are from the previous iteration. Since $c^l \lambda_i$ is also a constant, we can replace them by a normalization constant $C_i$ to get the final update equation:

$$p_i^\star(x_i) = p_i(x_i)\frac{\sum_{j \in Ne(i)}\sum_{x_j}\theta_{ij}(x_i,x_j)p_j(x_j)}{C_i}$$

The message-passing process for this version of CCCP is exactly the same as that for MAP QP and convex QP. This version does not require an inner loop as all the node marginals remain positive using such updates. This update process is also identical to the recently developed message-passing algorithm for MAP estimation that is based on expectation-maximization (EM) rather than CCCP [Kumar and Zilberstein, 2010]. However, CCCP provides a more flexible framework in that it handled the nonconvex and convex QP in a similar way as shown earlier. Furthermore, the CCCP framework allows for additional constraints to be added to the convex QP to make it tighter [Sriperumbudur and Lanckriet, 2009].

In sum, we have shown that the concave-convex procedure provides a unifying framework for the various quadratic programming formulations of the MAP problem. Each iteration of CCCP can be easily implemented using graph-based message passing. Interestingly, the messages exchanged for all the QP formulations we discussed remain exactly the same; the differences lie in how new node marginals are computed using such messages.

## 3 EXPERIMENTS

We now present an empirical evaluation of the CCCP algorithms. We first report results on synthetic graphs generated using the Ising model from statistical physics [Baxter, 1982]. We compare max-product (MP) and the CCCP message-passing algorithm for the QP formulation of MAP. We generated 2D nearest neighbor grid graphs for a number of grid sizes (ranging between $10 \times 10$ to $50 \times 50$) and varying values of the coupling parameter. All the variables were binary. The node potentials were generated by sampling from the uniform distribution $\mathcal{U}[-0.05, 0.05]$. The coupling strength, $d_{coup}$, for each edge was sampled from $\mathcal{U}[-\beta, \beta]$ following the mixed Ising model. The binary edge potential $\theta_{ij}$ was defined as follows:

$$\theta_{ij}(x_i,x_j) = \begin{cases} d_{coup} & x_i = x_j \\ -d_{coup} & x_i \neq x_j \end{cases}$$

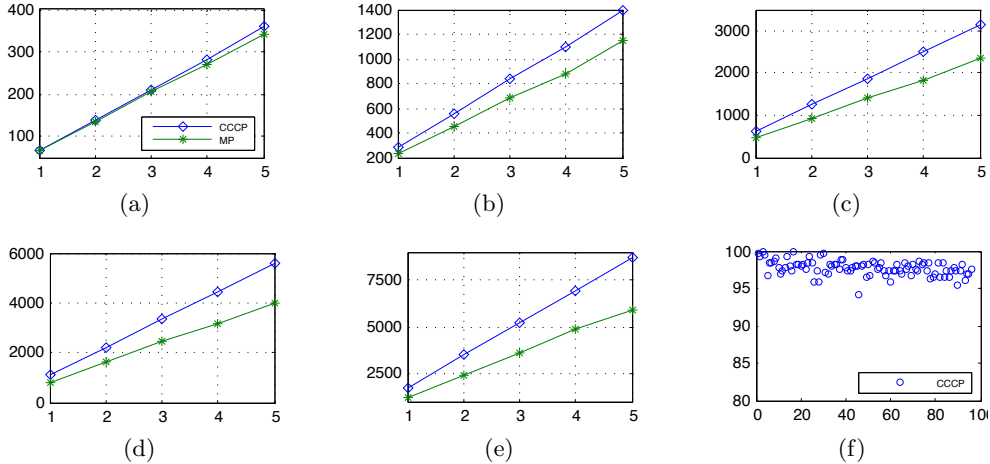

Figure 1: (a)–(e) show quality comparison between max-product (MP) and CCCP for Ising graphs with varying number of nodes (100–2500). The x-axis denotes the coupling parameter $\beta$, y-axis shows solution quality. (f) shows the solution quality CCCP achieves as a percentage of the optimal value (y-axis) for different protein design instances (x-axis).

For every grid size and each setting of the coupling strength parameter $\beta$, we generated 10 instances by sampling $d_{coup}$ per edge. For each instance, we considered the best solution quality of 10 runs for both max-product and CCCP. We then report the average quality of the 10 instances achieved for each parameter $\beta$. Both max-product and CCCP were implemented in JAVA and ran on a 2.4GHz CPU. Max-product was allowed 1000 iterations and often did not converge, whereas CCCP converged within 500 iterations.

Fig. 1(a–e) show solution quality comparisons between MP and CCCP. For $10 \times 10$ graphs (Fig. 1(a)), both CCCP and MP achieve similar solution quality. The gain in quality provided by CCCP increases with the size of the grid graph. For $20 \times 20$ grids, the average gain in solution quality, $((Q_{CCCP} - Q_{MP})/Q_{MP})$, for each coupling strength parameter $\beta$ is over 20%. For $30 \times 30$ (Fig. 1(c)) grids, the gain is above 30% for each parameter $\beta$; for $40 \times 40$ grids it is 36% and for $50 \times 50$ grids it is 43%. Overall, CCCP provides much better performance than max-product over these Ising graphs. And unlike max-product, CCCP monotonically increases solution quality and is guaranteed to converge. A detailed performance evaluation of the convex QP is provided [Ravikumar and Lafferty, 2006]. As such Ising graphs have relatively small QP representation, the CCCP message passing method and CPLEX had similar runtime for the convex QP.

We also experimented on the protein design benchmark (total of 97 instances) [Yanover et al., 2006]. In these problems, the task is to find a sequence of amino acids that is as stable as possible for a given backbone structure of protein. This problem can be modeled using a pairwise Markov random field. These problems are particularly hard and dense with up to 170 variables, each with a large domain size of up to 150 values. Fig. 1(f) shows the % of the optimal value CCCP achieves against the best upper bound provided by the LP based approach MPLP [Sontag et al., 2008]. MPLP has been shown to be very effective in solving exactly the MAP problem for several real-world problems. However for these protein design problems, due to the large variable domains, its reported mean running time is 9.7 hours [Sontag et al., 2008]. As Fig. 1(f) shows, CCCP achieves near-optimal solutions, on average within 97.7% of the optimal value. A significant advantage of CCCP is its speed: it converges within 1200 iterations for all these problems and requires $\approx 403$ seconds for the largest instance, much faster than MPLP. The mean running time of CCCP was $\approx 170$ seconds for this dataset. Thus CCCP can prove to be quite effective when fast, approximate solutions are desired. The main reason for this speedup is that CCCP's messages are easier to compute than MPLP's as also highlighted in [Kumar and Zilberstein, 2010]. Compared to the EM approach of [Kumar and Zilberstein, 2010], CCCP provides better solution quality: EM achieved 95% of the optimal value on average, while CCCP achieves 97.7%. The overhead of the inner loop in CCCP is small against EM which takes $\approx 352$ seconds for the largest instance, while CCCP takes $\approx 403$ seconds.

We also tested CCCP on the protein prediction dataset [Yanover et al., 2006]. The problems in this dataset are much smaller than those in the protein design dataset, and both max-product and MPLP achieve good solution quality. CCCP's performance was worse in this case, partly due to the local optima present in the nonconvex QP formulation of MAP. The convex QP formulation was not tight in this case.

Fig. 2(a) shows runtime comparison of CCCP against CPLEX for the convex QP for the 25 largest protein design problems w.r.t. the number of graph edges. After trying different QP solver options available in CPLEX, we chose the barrier method which provided the best performance. As CPLEX was quite slow, we let CPLEX use 8 CPU cores with 8GB RAM, while CCCP only used a single CPU. As this figure shows, CCCP is more than an order-of-magnitude faster than CPLEX even when it uses a single core. The longest CPLEX took was 3504 seconds, whereas CCCP only took 99 seconds for the same instance. The mean running time of CPLEX was 1914 seconds; for CCCP, it was 96 seconds. Surprisingly, CCCP converges in only 15 iterations to the optimal solution for all 25 problems. Fig. 2(b) shows the signed quality difference between CCCP and CPLEX for the convex QP objective. CPLEX provides the optimal solution within some non-zero $\epsilon$ (we used the default setting). This figure shows that even within 15 iterations, CCCP achieved a slightly better solution. The decoded solution quality provided by the convex QP was *decent*, within 80% of the optimal value, but not as high as the CCCP method for the nonconvex QP.

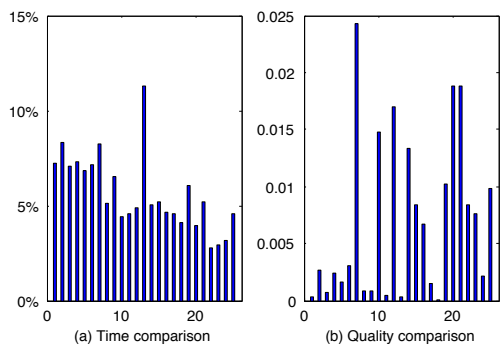

Figure 2: (a) Time comparison of CCCP for convex QP against CPLEX for the largest 25 protein design instances (x-axis). The y-axis denotes $T_{CCCP}/T_{CPLEX}$ as a percentage. (b) denotes the signed quality difference $Q_{CCCP} - Q_{CPLEX}$, a higher value is better.

## 4 CONCLUSION

We presented new message-passing algorithms for various quadratic programming formulations of the MAP problem. We showed that the concave-convex procedure provides a unifying framework for different QP formulations of the MAP problem represented as a *difference of convex functions*. The resulting algorithms were shown to be convergent – to a local optimum for the nonconvex QP and to the global optimum of the convex QP. Empirically, the CCCP algorithm was shown to work well on Ising graphs and real-world protein design problems. The CCCP approach provided much better solution quality than max-product for Ising graphs and converged significantly faster than max-product LP for protein design problems, while providing near optimal solutions. For the convex QP relaxation, CCCP provided more than an order-of-magnitude speedup over the state-of-the-art QP solver CPLEX. These results offer a powerful new way for solving efficiently large MAP estimation problems.

## Acknowledgments

Support for this work was provided in part by the NSF Grant IIS-0812149 and by the AFOSR Grant FA9550-08-1-0181.